# Sentiment Analysis of Covid-19 Tweets using Evolutionary Classification-Based LSTM Model

Arunava Kumar Chakraborty[1,*], Sourav Das[2], Anup Kumar Kolya[1]

[1] Dept. of Computer Science & Engineering, RCC Institute of Information Technology, Beleghata, Kolkata - 700015, India
[2] Maulana Abul Kalam Azad University of Technology, WB, Salt Lake, Kolkata - 700064, India
{arunava95.akc, sourav.das.research, anup.kolya}@gmail.com

**Abstract.** As the Covid-19 outbreaks rapidly all over the world day by day and also affects the lives of million, a number of countries declared complete lockdown to check its intensity. During this lockdown period, social media platforms have played an important role to spread information about this pandemic across the world, as people used to express their feelings through the social networks. Considering this catastrophic situation, we developed an experimental approach to analyze the reactions of people on Twitter taking into account the popular words either directly or indirectly based on this pandemic. This paper represents the sentiment analysis on collected large number of tweets on Coronavirus or Covid-19. At first, we analyze the trend of public sentiment on the topics related to Covid-19 epidemic using an evolutionary classification followed by the n-gram analysis. Then we calculated the sentiment ratings on collected tweet based on their class. Finally, we trained the long-short term network using two types of rated tweets to predict sentiment on Covid-19 data and obtained an overall accuracy of 84.46%.

## 1 Introduction

On 31st December, 2019 the Covid-19 outbreak was first reported in the Wuhan, Hubei Province, China and it started spreading rapidly all over the world. Finally, WHO announced Covid-19 outbreak as pandemic on 11th March, 2020, when the virus continues to spread [1]. Starting from China, this virus infected and killed thousands of people from Italy, Spain, USA, UK, Brazil, Russia, and other many more countries as well. On 21st August 2020, more than 22.5 million cases of Covid-19 were reported in more than 188 countries and territories, yielding more than 7,92,000 deaths; although 14.4 million people have reported to be recovered[1]. While this pandemic has continued to affect the lives of millions, many countries had enforced a strict lockdown for different periods to break the chain of this pandemic [1]. Since the Covid-19 vaccines are still yet to be discovered, therefore maintaining social distancing is the one and only one solution to check the spreading rate of this virus [2]. During the lockdown period a lot of people have chosen the Twitter to share their expression about this disease so we have been inspired to measure the human sensations about this epidemic by analyzing this huge Twitter data [3].

---

[1] https://gisanddata.maps.arcgis.com/apps/opsdashboard/index.html#/bda7594740fd40299423467b48e9ecf6

Initially, we have to face many challenges at the time of streaming the English tweets from the multilingual tweets all over the world as most of the peoples of foreign countries have used their native languages rather than English to express their feelings on social media [3]. However, we have developed our dataset considering the English tweets exclusively on Covid-19 of 160k tweets during April–May, 2020. We found the most popular words from the word corpus. Then we analyzed the trend of tweets using n-gram model. Further we assigned sentiment scores to our preprocessed tweets based on their sentiment polarity and classified our dataset on basis of their sentiment scores. Finally, we used those tweets and their sentiment ratings to train our LSTM model.

The following sections are furnished as follows: In Section 2 we have described some previous related research works. The architecture of our dataset and proposed pre-processing approach presented in Section 3. The Section 4 consists of Feature A for identifying the Covid-19 specified words based on the word lexicon. In Section 5, we have described Feature B as the trend of tweet words using n-gram model. The evolutionary classification on the sentiment-rated tweets based on their sentiment polarity has given in Section 6. In Section 7 we trained our LSTM model based on the classified tweets including their sentiment ratings. Whereas the Section 8 concludes the future prospects of our research work.

## 2 Related Works

A machine learning and cloud computing-based Covid-19 prediction model has been developed on May, 2020 to predict the future trend of this epidemic. They mainly used probabilistic distribution functions like Gaussian, Beta, Fisher-Tippet, and Log Normal functions to predict the trend [1].

A Covid-19 trend prediction model has introduced on June, 2020 for predicting the number of COVID-19 positive cases in different states of India. The researchers mainly developed a LSTM-based prediction model as LSTM model performs better for time series predictions. They tested different LSTM variants such as stacked, convolutional, and Bidirectional LSTM on the historical data, and based on the absolute error they found that Bi-LSTM gives more accurate results over other LSTM models for short-term prediction [4].

On July, 2020 the evolutionary K-means clustering on twitter data related to Covid-19 has been done by some researchers. They analyzed the tweet patterns using n-gram model. As the result they observed the difference between the occurrences of n-grams from the dataset [5].

Another research work describes a deep LSTM architecture for Message-level and Topic-based sentiment analysis. The authors used LSTM networks augmented with two kinds of attention mechanisms, on top of word embeddings pre-trained on a big collection of Twitter messages [7].

A group of researchers developed LSTM hyperparameter optimization for neural network-based Emotion Recognition framework. In their experiment they found that optimizing LSTM hyperparameters significantly improve the recognition rate of four-quadrant dimensional emotions with a 14% increase in accuracy and the model based on optimized LSTM classifier achieved 77.68% accuracy by using the Differential Evolution algorithm [8].

## 3 Preparing Covid-19 Dataset

Since this Covid-19 epidemic has affected the entire world, we have collected world-wide Covid-19 related English tweets at a rate of 10k per day in between April 19 and May 20, 2020 to create our dataset of about 160k tweets. The dataset we developed contains the important information about most of the tweets as its attribute. The attributes of our dataset are id [Number], created_at [DateTime], source [Text], original_text [Text], favorite_count [Number], retweet_count [Number], original_author [Text], hashtags [Text], user_mentions [Text], place [Text]. Finally, we have collected 1,61,400 tweets containing the hash-tagged keywords like—*#covid-19*, *#coronavirus*, *#covid*, *#covaccine*, *#lockdown*, *#homequarantine*, *#quarantinecenter*, *#socialdistancing*, *#stayhome*, *#staysafe* etc. In Fig. 1 we have represented an overview of our dataset.

| id | created_at | source | original_text | lang | favorite_count | retweet_count | original_author | hashtags | user_mentions | place |
|---|---|---|---|---|---|---|---|---|---|---|
| 125193476721131 | Sun Apr 19 | <a href="http: | RT @Ash_The | en | 0 | 705 | EmpoweringGo | Jehanaba | Ash_TheLoneW | Panjim Goa India |
| 125193473331715 | Sun Apr 19 | <a href="http: | RT @NGvision | en | 0 | 2646 | Ibilola_Amao | lockdown | NGvision2020 | London, England |
| 125193466618305 | Sun Apr 19 | <a href="http: | RT @Barnes_ | en | 0 | 5593 | cliff_skidmore | | Barnes_Law | Texas, USA |
| 125193460755922 | Sun Apr 19 | <a href="http: | RT @Joelpatri | en | 0 | 108 | GMA4Trump_ | Covid_19 | Joelpatrick1776 | Choctaw, OK |
| 125193458229277 | Sun Apr 19 | <a href="http: | RT @GlblCtzn | en | 0 | 4 | Macgirl730 | Together, | GlblCtzn | Menomonee Falls, WI |

**Fig. 1.** Partial snapshot of the Covid-19 tweets corpus.

### 3.1 Data Pre-Processing

Data pre-processing is mainly used for cleaning the raw data by following certain steps to achieve the better result for further evaluations. We have done the pre-processing on our collected data by developing a user defined pre-processing function based on NLTK (Natural Language Toolkit, a Python library for NLP). Stemming helps to reduce inflected words to their word stem, base, or root form whereas by using Tokenization this function splits each of the sentence into smaller parts of word.

# 4 Feature A: Covid-19 Specified Words Identification

After pre-processing we have developed the Bag-of-Words (BOW) model using the frequently occurred words from the word lexicon and we obtained a list of most frequent Covid-19 exclusive words. We have represented a dense word-cloud in Fig. 2 of some of the mostly used words within the corpus.

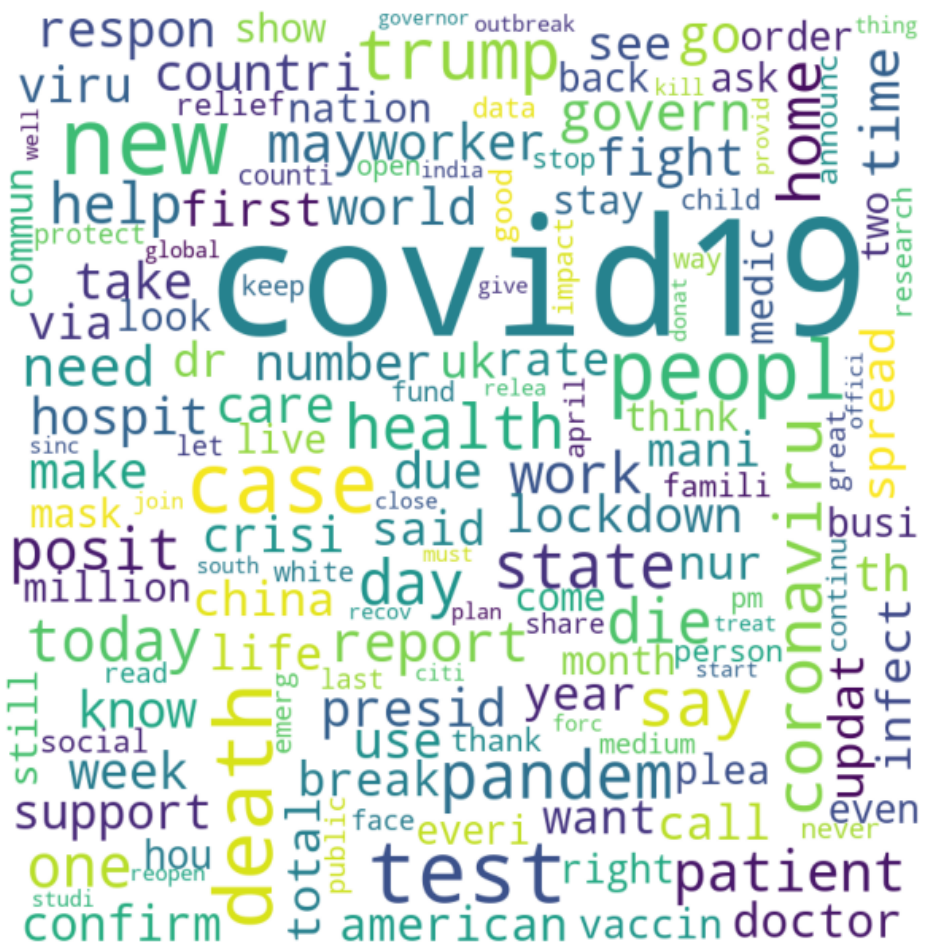

**Fig. 2.** Some of the most popular Covid-19 related words from our corpus.

## 4.1 Word Popularity

Several words within the generated corpus have been found at different times in different positions of the tweets. Here we have counted the recurrence of each word and presented the top 50 popular words along with their popularity in Fig. 3.

After finding the word popularity, we have calculated the probability of repetition for each word on the basis of total 3,53,704 words from the corpus. Table 1 represents the popularity and probability scores of some most frequent words.

$$P(W_i) = \frac{count(W_i)}{\sum_{i=0}^{n} count(W_{i=0}^{n})} \quad (1)$$

**Table 1.** Popularity & probability of most frequent words.

| | Words | Popularity | Probability |
|---|---|---|---|
| 0 | Covid19 | 91794 | 0.259522 |
| 1 | Test | 11663 | 0.032974 |
| 2 | New | 11305 | 0.031962 |
| 3 | People | 10834 | 0.030630 |
| 4 | Death | 10783 | 0.030486 |

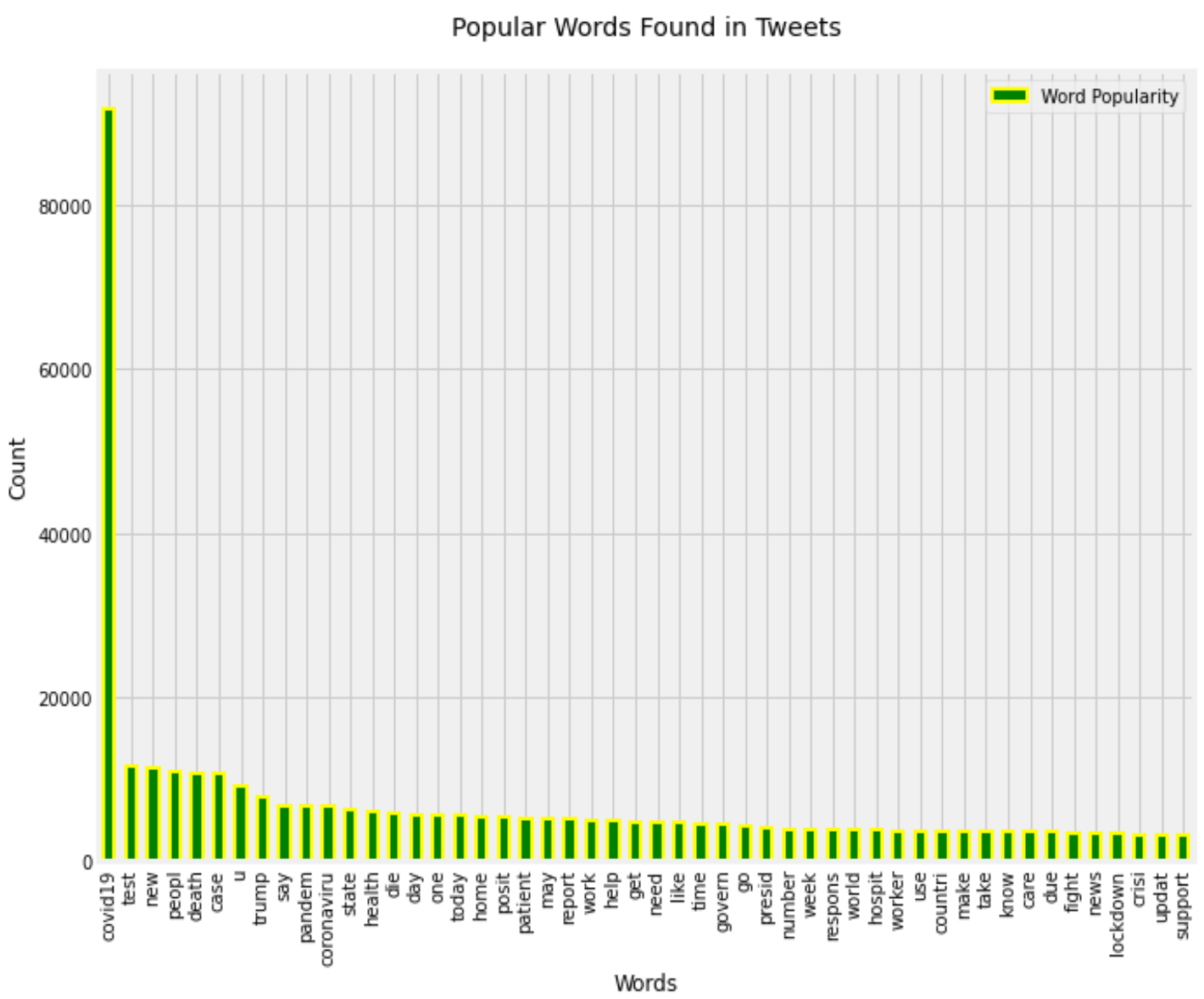

**Fig. 3.** Graphical representation of the popularity for most frequent Covid-19 exclusive words.

## 5 Feature b: Word Popularity Detection using N-gram

Lexical n-gram models are widely used in Natural Language Processing for statistical analysis & syntax feature mapping. We developed n-gram model to analyze our generated corpus consisting of tokenized words for finding the popularity of words or group of adjacent words. Here the probability of the occurrence of a sequence can be calculated using probability chain rule:

$$P(x_1, x_2, x_3, \ldots x_n) = P(x_1)\, P(x_2 \mid x_1)\, P(x_3 \mid x_1, x_2) \ldots P(x_n \mid x_1, x_2, x_3, \ldots x_{(n-1)}) \qquad (2)$$

$$= \prod_{i=1}^{n} P(x_i \mid x_1^{(i-1)}) \qquad (3)$$

For example, we can consider a sentence as “Still Covid-19 wave is running”. Now as per the probability chain rule, P(“Still Covid19 wave is running”) = P(“Still”) x P(“Covid19” | “Still”) x P(“wave” | “Still Covid19”) x P(“is” | “Still Covid19 wave”) x P(“running” | “Still Covid19 wave is”).

The probabilities of words in each sentence after applying probability chain rule:

$$P(W_1, W_2, W_3, \ldots W_n) = \prod_j P(W_j \mid W_1, W_2, W_3, \ldots W_{(j-1)}) \qquad (4)$$

$$= \prod_{j=1}^{n} P(W_j \mid W_1^{(j-1)}) \qquad (5)$$

The bigram model estimates the probability of a word by using only the conditional probability $P(W_i|W_{i-1})$ of one preceding word on given condition of all the previous words $P(W_i|W_1^{i-1})$ [6].

$$P(W_1, W_2) = \prod_{i=2} P(W_2 \mid W_1) \quad (6)$$

The expression for the probability is -

$$P(W_k \mid W_{(k-1)}) = \frac{\text{count}\,(W_{(k-1)}, W_k)}{\text{count}\,(W_{(k-1)})} \quad (7)$$

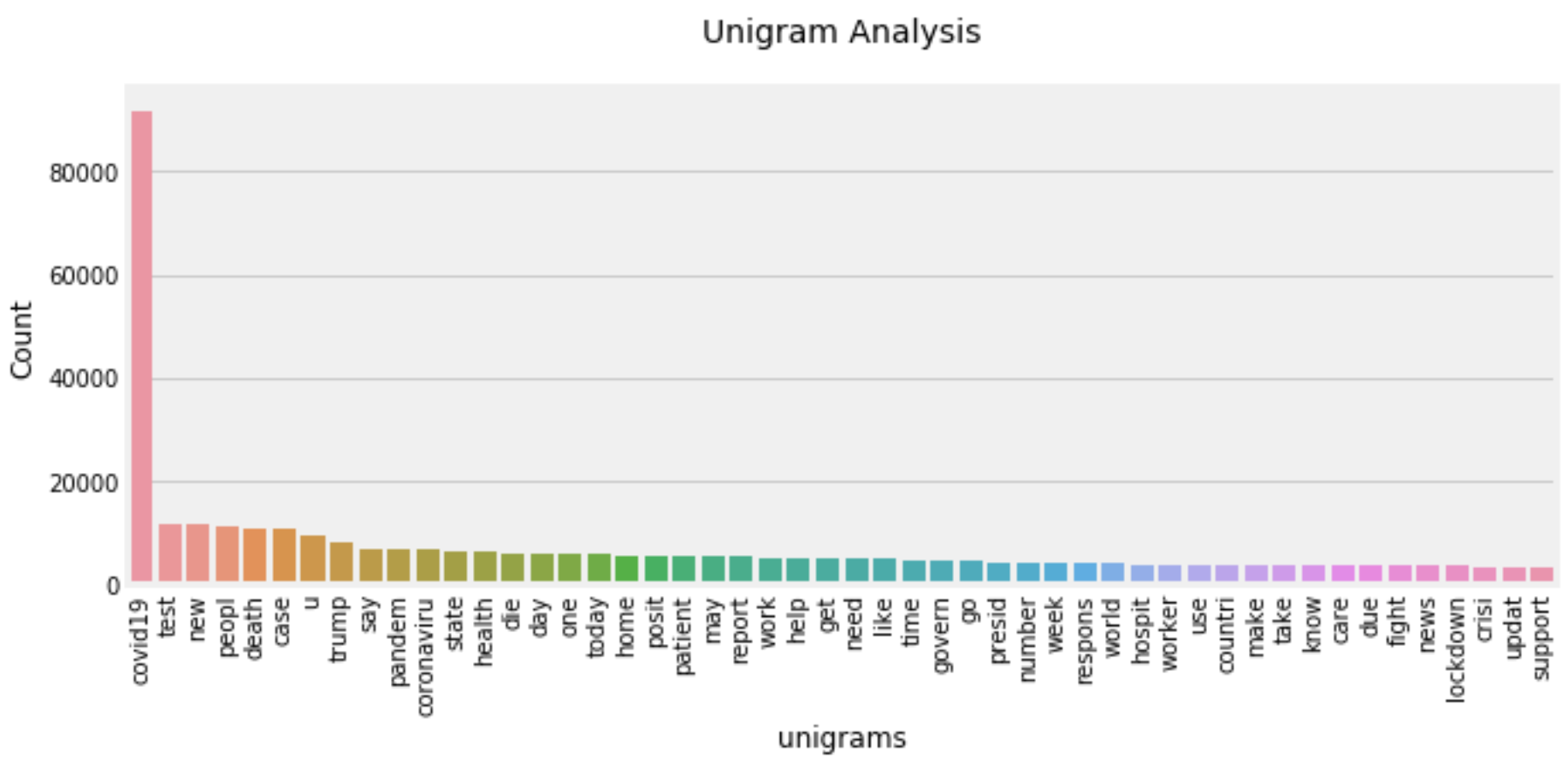

**Fig. 4.** Graphical representation of the popularity for most frequent unigrams.

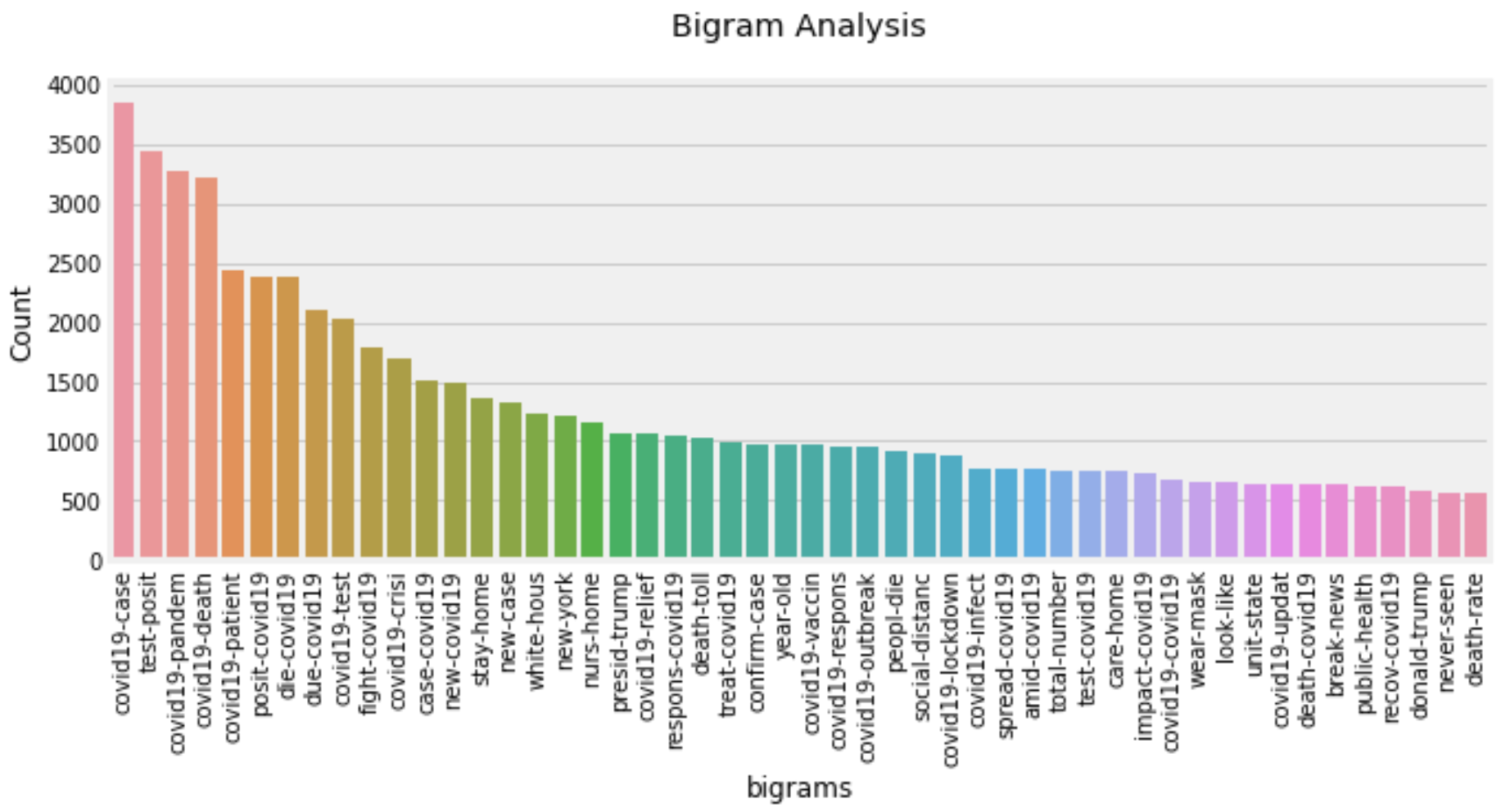

**Fig. 5.** Graphical representation of the popularity for most frequent bigrams.

We have identified the most popular unigrams, bigrams, and trigrams within our corpus using the n-gram model. The graphical representations of 50 most popular unigrams, bigrams, and trigrams along with their popularity are presented in Fig. 4, Fig. 5 and Fig. 6 respectively.

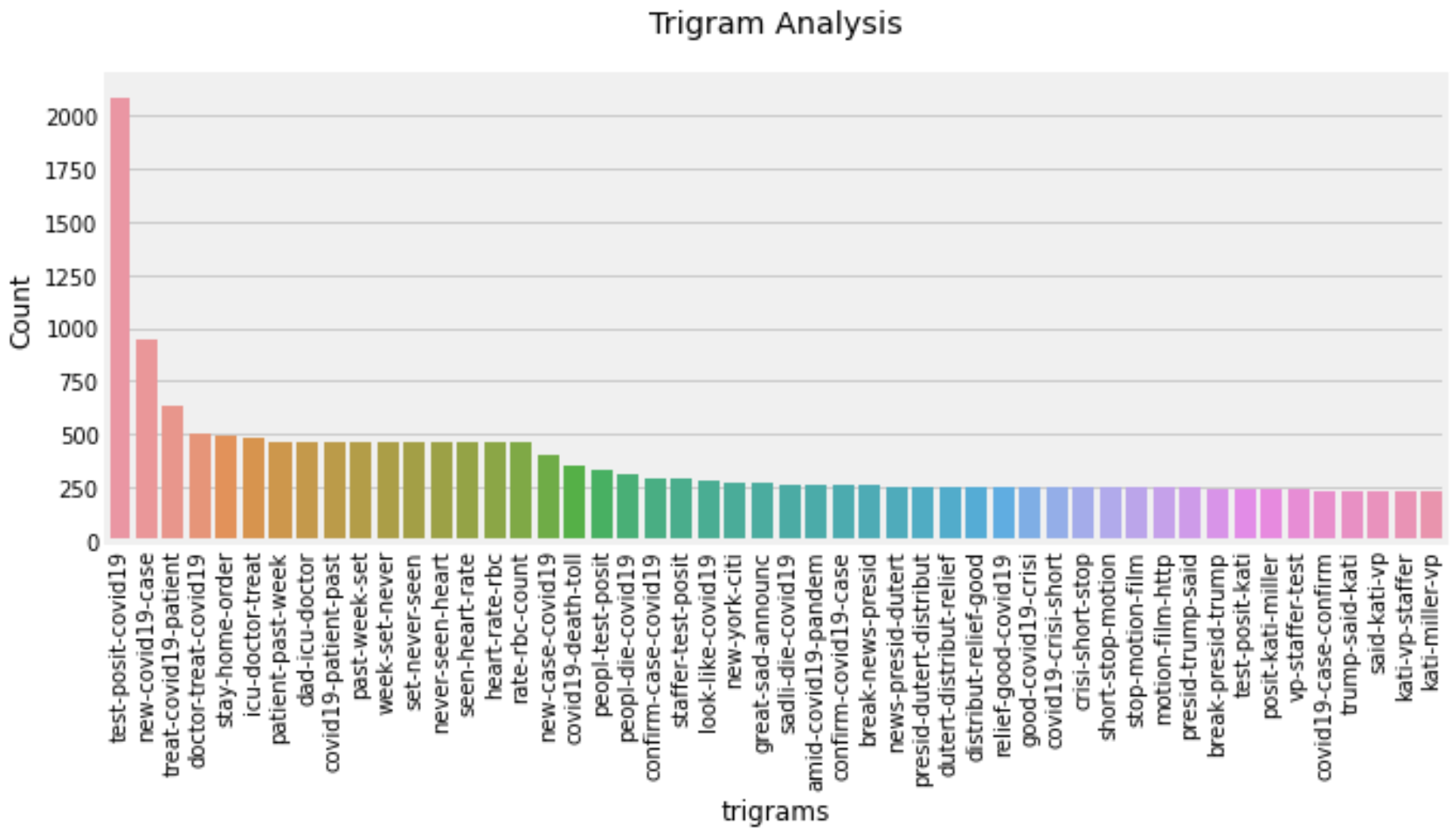

**Fig. 6.** Graphical representation of the popularity for most frequent trigrams.

As the result of this analysis, we found that the popularity of trigrams is lesser than that of bigrams and the unigrams popularity is the highest according to this n-gram model.

# 6 Sentiment Analysis

To measure the trend of public opinions we use Sentiment Analysis, a specific type of Data Mining through Natural Language Processing (NLP), computational linguistics and text analysis. The subjective information from the social media are analyzed and extracted to classify the text in multiple classes like positive, negative, and neutral.

Here we calculated the sentiment polarity of each cleaned and preprocessed tweet using the NLTK-based Sentiment Analyzer and get the sentiment scores for *positive*, *negative* and *neutral* classes to calculate the *compound* sentiment score for each tweet.

## 6.1 Sentiment Classification

We have classified the tweets on the basis of the *compound* sentiments into three different classes, i.e. *Positive*, *Negative* and *Neutral*. Then we have assigned the sentiment polarity rating for each tweet based on the algorithm presented in Table 2.

**Table 2.** Algorithm used for sentiment classification of our Covid-19 tweets.

| Algorithm Sentiment Classification of Tweets (compound, sentiment): |
|---|
| *for each k in range (0, len(tweet.index)):* |
|     *if* $tweet_k$*[compound] < 0:* |
|         $tweet_k$*[sentiment] = 0.0 # assigned 0.0 for Negative Tweets* |
|     *elif* $tweet_k$*[compound] > 0:* |
|         $tweet_k$*[sentiment] = 1.0 # assigned 1.0 for Positive Tweets* |
|     *else:* |
|         $tweet_k$*[sentiment] = 0.5 # assigned 0.5 for Neutral Tweets* |
| *end* |

In Fig. 7, we have represented the sentiment classification with the overall percentage of each positive, negative, and neutral tweet found in the dataset. It can be visualized that the sentiment classes are naturally imbalanced as a large portion of social media users are either negative or neutral against the Covid-19 and the medical details associated with it.

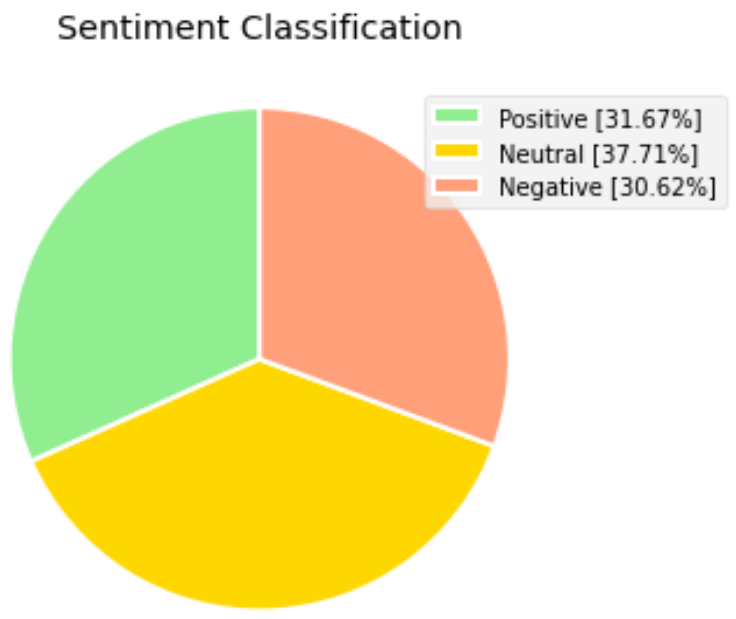

**Fig. 7.** Sentiment distribution of three class polarity along with the percentage of Covid-19 tweets occurred from each class.

## 7 Sentiment Modeling using Sequential LSTM

In traditional textual sentiment analysis, LSTM (Long-Short Term Model) network have already been proven to be performing better than the similar neural models [3]. We exploit a sequential LSTM model for sentiment evaluation of our Covid-19 dataset. We developed a new dataset consisting of the cleaned and preprocessed tweets along with their corresponding *positive* (1.0) and *negative* (0.0) sentiments. Then we created two sets *X* and *y* for the cleaned tweets and their sentiment scores, respectively and split the dataset into 80:20 ratio, i.e., 80% for training (*X_train*, *y_train*) and 20% for validation (*X_test*, *y_test*) purposes, respectively. A large number of Covid-19 exclusive words were generated by this model from the new dataset. Then we converted these words into word vectors using *word2vec* by setting the vector dimension as 200 for each collected n-grams within a sentence and developed new *X_train*, *X_test* sets consisting with the calculated word vectors for further processing. From updated training set the word vectors and the respective sentiment scores fed in the model as the first layer of inputs. In this experiment, we used *TensorFlow* framework and *keras* library to add *Sequential* LSTM model with *Dense* layers. We have trained our five-layered model for 30 epochs with two types of outline activation function with parameters, optimizer, loss, and accuracy. We have used *ReLU* (Rectified Linear Unit) activation function for the initial set of *Dense* layers with 128, 64, and 32 units, respectively and *Sigmoid* activation function for the outermost final *Dense* layer with 2 units. During the training we have used 32 batches and 2 verbose for our model. For some epochs Table 3 represents the training accuracy vs. loss and validation accuracy vs. loss, respectively.

**Table 3.** Training accuracy vs. loss, validation accuracy vs. loss within 30 epochs.

| Epochs | Train Loss | Train Accuracy | Val Loss | Val Accuracy |
|---|---|---|---|---|
| Initially | 59.93% | 67.63% | 56.18% | 70.45% |
| $5^{th}$ | 42.71% | 79.27% | 43.75% | 78.74% |
| $10^{th}$ | 35.91% | 83.38% | 39.58% | 81.71% |
| $15^{th}$ | 30.43% | 86.36% | 39.28% | 82.81% |
| $20^{th}$ | 26.23% | 88.40% | 41.04% | 83.47% |
| $25^{th}$ | 22.44% | 90.15% | 41.96% | 84.24% |
| $30^{th}$ | 19.38% | **91.67%** | 45.57% | **84.46%** |

After completion of the training on our model, we have finally achieved 91.67% of overall training accuracy whereas the validation accuracy is 84.46% on the testing data. Table 4 and Table 5 are representing the confusion matrix and classification report to present the differences between predicted and the actual tweets along with the different classes.

**Table 4.** Confusion Matrix.

| **Actual** | | **Predicted** | |
|---|---|---|---|
| | | **Positive** | **Negative** |
| | Positive | 8298 (TP) | 1941 (FP) |
| | Negative | 1946 (FN) | 7924 (TN) |

**Table 5.** Classification Report.

| | Precision | Recall | F1-Score | Support |
|---|---|---|---|---|
| Positive (1.0) | 0.81 | 0.81 | 0.81 | 10239 |
| Negative (0.0) | 0.80 | 0.80 | 0.80 | 9870 |
| Avg / Total | 0.81 | 0.81 | 0.81 | 20109 |

In Fig. 8, we have plotted the percentages of training accuracy vs. loss and validation accuracy vs. loss, achieved by the Sequential LSTM model during compilation. From the figure it is evident that there has been a significant loss difference between the training and testing epochs. This indicates a slight overfitting of the data which can be postulated from the several tweet collection parameters differing from time to time in the tweets streaming phase.

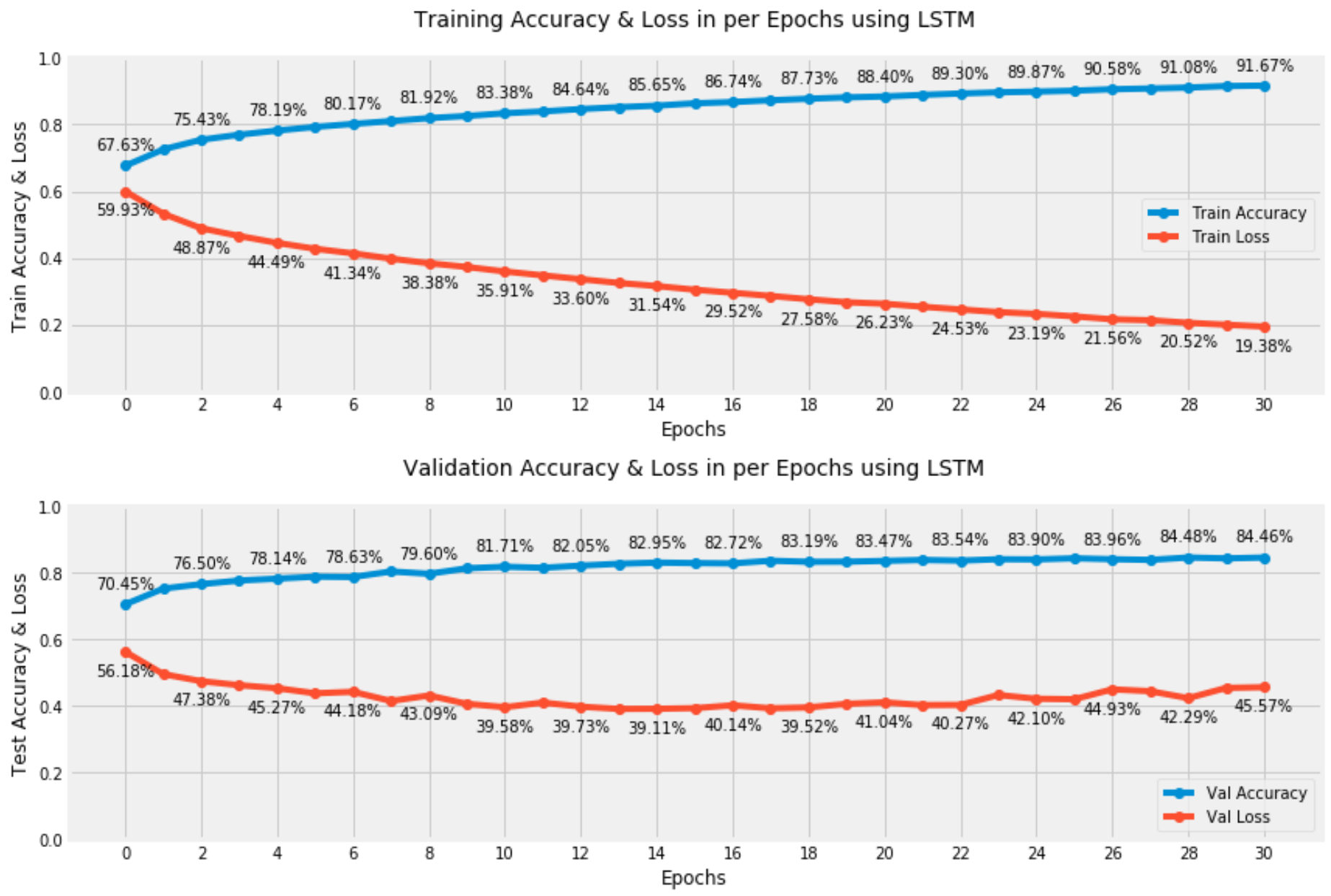

**Fig. 8.** Performance metrics from the training loss vs. accuracy and validation loss vs. accuracy by the proposed model.

# 8 Conclusion & Future Scope

This experiment is mainly focused on Deep Learning-based Sentiment Analysis on Covid-19 tweets. We extracted the mostly popular words and analyzed the popularity of group of words using n-gram model as two main features of our dataset. However, later we developed a model to assign the sentiment ratings to the tweets based on their sentiment polarities calculated by sentiment analyzer and classify all tweets into *positive* and *negative* classes based on their assigned sentiment ratings. Then, using this classified dataset containing the cleaned and preprocessed tweets and their sentiment ratings, i.e., 1.0 for *positive* and 0.0 for *negative*, we trained our Deep Learning-based LSTM model. We divided the dataset into 80:20 ratio, i.e., 80% for training and 20% for testing purposes. After running 30 epochs on almost 93,474 parameters, we achieved validation accuracy as 84.46%.

For the future work, we want to develop a polarity-popularity model based on the features extracted during this experiment so that we can assign the refined sentiment ratings to the tweet based on the polarity of mostly recurred words [3]. With that data we will train the deep learning model to enhance the validation accuracy of our system.